\documentclass[11pt,a4paper]{article}
\usepackage[hyperref]{acl2018}
\usepackage{times}
\usepackage{latexsym}

\usepackage{amsmath}

\usepackage[]{graphicx}

\usepackage{setspace}

\usepackage{url}
\usepackage{color}

\usepackage{amsmath}

\usepackage[skip=4pt]{caption}

\usepackage{titlesec}
\titlespacing*{\section}
{0pt}{1.3ex plus 1ex minus .2ex}{1.3ex plus .2ex}
\titlespacing*{\subsection}
{0pt}{1.0ex plus 1ex minus .2ex}{1.0ex plus .2ex}
\titlespacing*{\subsubsection}
{0pt}{1.0ex plus 1ex minus .2ex}{1.0ex plus .2ex}

\aclfinalcopy 

\title{Knowledge-based end-to-end memory networks}

\author{Jatin Ganhotra \\
  IBM Research \\
  {\tt jatinganhotra@us.ibm.com} \\\And
  Lazaros Polymenakos \\
  IBM Research \\
  {\tt lcpolyme@us.ibm.com} \\}

\date{}

\begin{document}

\setlength{\abovedisplayskip}{0.05cm}
\setlength{\belowdisplayskip}{0.05cm}

\maketitle
\begin{abstract}


End-to-end dialog systems have become very popular because they hold the promise of learning directly from human to human dialog interaction. Retrieval and Generative methods have been explored in this area with mixed results. A key element that is missing so far, is the incorporation of a-priori knowledge about the task at hand. This knowledge may exist in the form of structured or unstructured information. As a first step towards this direction, we present a novel approach, Knowledge based end-to-end memory networks (KB-memN2N), which allows special handling of named entities for goal-oriented dialog tasks. We present results on two datasets, DSTC6 challenge dataset and dialog bAbI tasks. 

\end{abstract}

\section{Introduction}
End-to-end dialog systems, based on neural networks have shown promising performance in non goal-oriented chit-chat settings (\citet{shang2015neural}; \citet{vinyals2015neural}; \citet{sordoni2015neural}; \citet{serban2016building}; \citet{dodge2015evaluating}), where they are trained to predict the next utterance in social media forums (\citet{ritter2011data}; \citet{wang2013dataset}; \citet{lowe2015ubuntu}) or movie conversations (\citet{banchs2012movie}). In end-to-end dialog systems, all components are directly trained on past dialogs and
have shown the promise of learning directly from human-to-human dialog interactions. Such approaches are either \textit{generative} (\citet{le2016lstm}, \citet{ghazvininejad2017knowledge},  \citet{goyal2016natural}), where the system generates the next utterance in the conversation word-by-word, or \textit{retrieval-based}, where the system has to pick the next utterance from a list of potential responses. 

But the performance achieved on chit-chat may not necessarily carry over to goal-oriented conversations. Additionally, these systems lack the incorporation of a-priori task-related knowledge. A-priori knowledge about the task comes from either structured (e.g. databases, websites) or unstructured data sources (e.g. documents). We show that incorporating this a-priori knowledge can help in improving the performance of goal-oriented end-to-end dialog systems.

\citet{bordes2016learning} evaluated end-to-end memory networks proposed by \citet{sukhbaatar2015end} for goal-oriented dialog bAbI tasks, where the memory component effectively acts as a (dynamic) representation of the dialog context and allows for inference and reasoning over it. The system achieves good performance on the dialog bAbI tasks and incorporates a-priori knowledge about entities from the knowledge base (KB) via match-type features (explained in Section \ref{Experiments}). But they treat the entities used in dialog as part of the vocabulary and use a standard bag-of-words encoding for utterances, where embeddings for all words(including entities) are added to generate the utterance embedding. For example, the utterance embedding for \textit{"i'd like to book a table in london"} is generated by adding the embeddings for all words (\textit{'i'd', 'like', 'to', 'book', 'a', 'table', 'in', 'london'}) together. Position encoding for sentences proposed by \citet{sukhbaatar2015end} tries to capture the order of words, but entities are still treated the same way as other words e.g. \textit{a}, \textit{in}, \textit{like} etc. 

We extend their model and propose Knowledge-based end-to-end memory networks, wherein we treat entities present in the KB separately. Each entity in the dialog is represented by separate individual memory and entity embeddings are not added to generate the sentence embedding. This allows our model to perform separate attention over the entities and identify the entities relevant for the next system utterance. For goal-oriented dialog systems, system utterances usually include entities encountered in the dialog so far. Therefore, our proposed handling of entities can help the model by retrieving the response with the correct entity in a retrieval setting, or include the correct entity in the generated response word-by-word in a generative setting. We demonstrate the new approach on two data sets related to the restaurant reservation use-case below.

\section{Datasets and Task description}
We perform experiments on two goal-oriented dialog datasets: dialog bAbI tasks from \citet{bordes2016learning} and Dialog System Technology Challenge (DSTC6) dataset (End-to-end goal-oriented dialog learning track) from \citet{perezdialog}.

Both datasets are based on restaurant reservation and provide a KB which contains information about restaurants such as their name, location, phone number, cuisine etc.. The model needs to predict the next system utterance, given a user utterance (query) and dialog history (story/context). The dialog history provided may contain information about restaurants from the KB for certain tasks mentioned below.

There are 5 goal-oriented tasks in bAbI dialog tasks - Task 1: Issuing API calls, Task 2: Updating API calls, Task 3: Displaying options, Task 4: Providing extra information and Task 5: Conducting full dialogs. An example dialog for Task 1 is mentioned below. Tasks 1 and 2 test if the model can implicitly track dialog state. The system must learn to ask questions to collect information about user preferences and generate an API call. Task 3 and 4 check if the model can use KB facts(entities) in a dialog setting. In task 3, the model must present to the user the restaurants (which matched his/her preferences) in the decreasing order of their rating, until the user accepts an option. Task 4 requires the model to answer factual questions from the user about restaurant information such as phone number or address. Task 5 combines all the previous tasks and provide a full dialog setting for training the model. In addition to KB, a global list of all possible system responses (candidates) for all tasks is provided and the system must predict the correct candidate for a given dialog. For bAbI dialog tasks, the dataset consists of 1000 dialogs each in training, validation and test. An example for each task is provided in Appendix A.

The DSTC6 challenge dataset from \citet{perezdialog} is similar to dialog bAbI tasks, but differs in the number of user preferences needed for a reservation and also has more varied user utterances. Each dialog is provided with 10 possible system responses (candidates) and the system must predict the correct candidate from the 10 candidates provided. For DSTC6 dataset, the dataset consists of 10,000 dialogs, split into 8000 dialogs for training and 1000 dialogs each for validation and test set\footnote{True labels for the test set for DSTC6 challenge dataset have not been released yet. Therefore, we split the dataset provided into training, validation and test sets for evaluation.}.

\section{Methods}
We extend the memory network architecture proposed by \cite{bordes2016learning}, where we treat entities (symbols present in KB) separately to identify the entities relevant for the next system utterance. The section below describes the original approach, our proposed solution and also explains why our approach is better suited for goal-oriented dialog tasks which involve a KB.

\subsection{End-to-end memory networks (memN2N)}
End-to-end memory Networks (\citet{sukhbaatar2015end}) are a recent class of models that have been applied to a range of natural language processing tasks. They use memory to store context and perform reasoning over it. The memory is updated iteratively using hops (multiple layers) and is used to reason the required response.

A single layer version of the model is described below in equation \ref{eq1}. A given sentence $(i)$ from the context (dialog history) is stored in the memory by it's input representation $(m_{i})$. Each sentence $(i)$ also has a corresponding output representation $(c_{i})$. The context $(m_{1}, ..., m_{i} ; c_{1}, ..., c_{i})$ and the query q$(u)$ are represented via embeddings learned for the vocabulary. To identify the relevance of a memory for the next-utterance prediction, attention of query over memory is computed via dot product, where $(p_{i})$ represents the probability for each memory in equation \ref{eq1}. An output vector $(o)$ is computed by a weighted sum of the memory embeddings $(c_{i})$ with their corresponding probabilities in equation \ref{eq2}. The output vector $(o)$ represents the overall embedding for the context. The output vector $(o)$ and query $(u)$ are then passed through a final weight matrix used $(W)$ and a softmax to produce the predicted label $\hat{a}$ in equation \ref{eq3}. Note that the sentence representation $(m_{i} ; c_{i})$ is generated using a bag-of-words encoding i.e. by adding embeddings for each word $(x_{1}, ..., x_{j})$ in the sentence \cite{sukhbaatar2015end}.
\begin{align}
\label{eq1}
p_{i} &= \textrm{Softmax}(u^{T} (m_{i})) \\
\label{eq2}
o &= \sum_{i}p_{i}c_{i} \\
\label{eq3}
\hat{a} &= \textrm{Softmax}(W (o + u))
\end{align}

\begin{figure*}[t]
\begin{center}
\includegraphics[width=\textwidth]{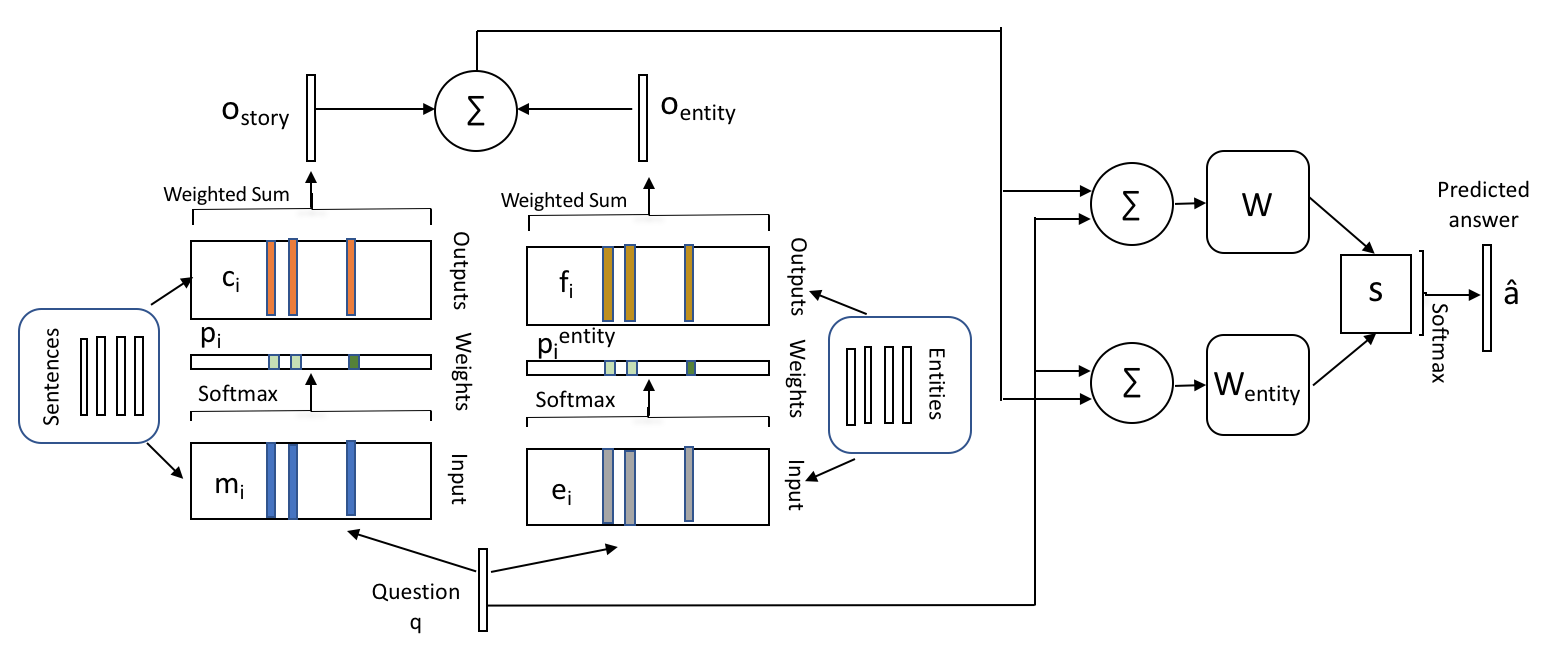}
\end{center}
\vspace{0pt}
\caption{Single layer version of Knowledge-based end-to-end memory networks}
\label{fig_kb_memn2n}
\end{figure*}

\subsection{Knowledge-based end-to-end memory networks (KB-memN2N)}
We propose Knowledge-based end-to-end memory networks, where we make the following changes. A single layer version of the model is shown in Fig. \ref{fig_kb_memn2n}.
\begin{enumerate}
\itemsep0em

    \item We replace each entity with it's entity-type token in the utterance. For example, \textit{"i'd like to book a table in london"} is changed to \textit{"i'd like to book a table in LOCATION"}. The entity-type tokens (\textit{LOCATION, CUISINE etc.}) are added to the vocabulary, where the model learns a generic embedding for each entity-type token.
    
    \item We represent entities present in the context via separate individual memories $(e_{i})$. For example, \textit{"may i have a table with spanish cuisine in rome"} is represented by $(m_{1})$ for \textit{"may i have a table with CUISINE in LOCATION"} and $(e_{1}, e_{2})$ for \textit{spanish} and \textit{rome} as shown in Fig.\ref{fig_kb_memn2n}. This allows us to capture and represent entities separately within the neural framework, so that they can be used for attention later.
    
    \item Dual-attention over story and entities: Since we have separate memory vectors for context $(m_{1}, ..., m_{i})$ and entities $(e_{1}, ..., e_{i})$, we perform separate attention over them to generate individual output vectors $(o_{story})$ and $(o_{entities})$ for context and entities as shown in equation \ref{eq6} and \ref{eq7}. The separate attention over entities allows us to find relevant entities that will be part of the system output response.
    
    \item Dual-attention over candidates and candidate entities: We also treat entities present in the candidates separately. We learn separate weight matrices $(W)$ and $(W_{entity})$ for candidates and the entities present in candidates as shown in Fig.\ref{fig_kb_memn2n}. The final prediction is made over both candidates and entities (equation \ref{eq9} and \ref{eq10}) from candidates and scores $(s_{story}, s_{entities})$ are merged together before a softmax operation is applied for final prediction $(\hat{a})$ as shown in equations \ref{eq11} and \ref{eq12}.
\end{enumerate}
\begin{align}
\label{eq4}
p_{i} & = \textrm{Softmax}(u^{T} (m_{i})) \\
\label{eq5}
p^{entity}_{i} & = \textrm{Softmax}(u^{T} (e_{i})) \\
\label{eq6}
o_{story} & = \sum_{i}p_{i}c_{i} \\
\label{eq7}
o_{entities} & = \sum_{i}p^{entity}_{i}f_{i} \\
\label{eq8}
o & = o_{story} + o_{entities} \\
\label{eq9}
s_{story} & = W (o + u) \\
\label{eq10}
s_{entities} & = W_{entity} (o + u) \\
\label{eq11}
s & = s_{story} + s_{entities} \\
\label{eq12}
\hat{a} & = \textrm{Softmax}(s))
\end{align}

\section{Experiments and Results} \label{Experiments}
We perform experiments on two goal-oriented dialog datasets: dialog bAbI tasks and DSTC6 dataset. During training, the predicted candidate $\hat{a}$ is used to minimise a standard cross-entropy loss with the true candidate ${a}$ against, the 10 candidates provided with each dialog for DSTC6 dataset and the global list of candidates for bAbI dialog tasks. We use Adam \cite{kingma2014adam} for optimization. We perform our experiments with the following hyper-parameter values: embedding dimension p = 30, learning rate $\lambda$ = 0.001 and number of hops K = 3. We use Per-response accuracy for evaluation of our models (per-response accuracy counts the percentage of responses that are correct).

Table \ref{table:dstc} shows the per-response accuracy of both models on the DSTC6 challenge dataset. We observe that our proposed idea achieves higher performance on all tasks, except task 2. The accuracy increases considerably on tasks 4 and 5, where task 5 represents the combination of all 4 tasks.

\begin{table}[t]
\centering
\begin{tabular}{c|c|c}
\hline
Task & memN2N & KB-memN2N \\ \hline
1	& 98        & \textbf{98.1} \\ \hline
2	& \textbf{100}    & 95.8 \\ \hline
3	& 96.4      & \textbf{97.1} \\ \hline
4	& 82.8     & \textbf{86.2} \\ \hline
5	& 90.1     & \textbf{93.5}\\ \hline
\end{tabular}
\caption{Mean accuracy \% on the DSTC6 dataset\footnote{We plan to perform experiments with match-type features in the future.}}
\label{table:dstc}
\end{table}

Table \ref{table:babi} shows the per-response accuracy of both models on bAbI dialog tasks. We observe that our model performs similar to the original memN2N architecture for tasks 1, 3 and 4. We observe reduction in accuracy for tasks 2 and 5 with respect to memN2N.

By investigation, we find that bAbI dialog dataset is simpler compared to the DSTC6 challenge dataset\footnote{\label{footnote1}For task1, the number of unique user utterances is 420 for bAbI dialog tasks and 1762 for DSTC6 dataset. We calculate unique user utterances by replacing entities with their entity-type tokens}. In addition to a smaller set of unique user utterances, the natural language used for user utterances is simpler and has less variations (uses a small set of templates) compared to the DSTC6 dataset. We believe that the simplicity of bAbI dialog dataset allows the original memN2N model to perform better. However, for DSTC6 dataset, our KB-memN2N performs better than memN2N, as shown in Table \ref{table:dstc}, because KB-memN2N allows dual attention over context and entities.

\begin{table}[t]
\centering
\begin{tabular}{c|p{1.6cm}|p{2cm}|p{1.8cm}}
\hline
Task & memN2N & memN2N (match-type) & KB-memN2N \\ \hline
1	& 99.9     & \textbf{100}    & \textbf{100} \\ \hline
2	& \textbf{100}   & 98.3   & 91.9 \\ \hline
3	& \textbf{74.9}     & \textbf{74.9}   & 74.8 \\ \hline
4	& 59.5     & \textbf{100}    & 57.2 \\ \hline
5	& \textbf{96.1}     & 93.4   & 92.8 \\ \hline
\end{tabular}
\caption{Mean accuracy \% on dialog bAbI tasks; Results for memN2N and memN2N (match-type) are taken from \citet{bordes2016learning}.}
\label{table:babi}
\end{table}

\citet{bordes2016learning} propose match-type features for handling entities. Results for the model including match-type features are shown in Table \ref{table:babi} under memN2N (match-type). For Match-type features, entity-type tokens (e.g. ADDRESS, PHONE etc.) are added to a candidate if an entity is present both in the candidate and the story for a given dialog. For example, for a task 4 dialog with restaurant information about \textit{RES\_ABC}, \emph{only} one candidate \textit{"here it is RES\_ABC\_address"} will be modified to \textit{"here it is RES\_ABC\_address ADDRESS"}. If the query (last user utterance) is for the restaurant's address, then using match-type features reduces the output search space and allows the model to attend to specific candidates better. With match-type features, the accuracy for the model increases for task 4 but decreases for tasks 2 and 5 (conducting full dialogs). Hence, match-type features can only work in a retrieval setting and will not work in a generative setting where the next system utterance is generated word-by-word. Our KB-memN2N model will work in both retrieval and generative settings.


\section{Conclusion and Future work}
In this paper we proposed a method for handling entities and perform empirical evaluation on two goal-oriented dialog datasets. Our experiments on the DSTC6 dataset illustrate that our proposed handling for entities performs better than models without special entity handling, especially for Task 5 (conducting full dialogs). In future, we plan to perform experiments on the Out-of-vocabulary test dataset and a combination of KB-memN2N with match-type features. We are also interested in testing the idea in a generative setting to show the benefits of our approach.

\newpage
\bibliography{acl2018}
\bibliographystyle{acl_natbib}

\newpage
\appendix
\gdef\thesection{Appendix \Alph{section}}
\section{\\bAbI dialog tasks' examples}
\label{appendix_a}

\textbf{dialog bAbI task 1:}
\begin{spacing}{0.9}
{\footnotesize
\noindent
\\
U: hello\\
\texttt{S: hello what can i help you with today}\\
U: can you book a table for six people with french food\\
\texttt{S: i'm on it}\\
\texttt{S: where should it be}\\
U: in bombay\\
\texttt{S: which price range are looking for}\\
U: i am looking for a cheap restaurant\\
\texttt{S: ok let me look into some options for you}\\
\texttt{S: api\_call french bombay six cheap}\\
}
\end{spacing}

\noindent\textbf{dialog bAbI task 2:}
\begin{spacing}{0.9}
{\footnotesize
\noindent
\\
U: hello\\
\texttt{S: hello what can i help you with today}\\
U: can you book a table in a expensive price range for eight people with british cuisine in london\\
\texttt{S: i'm on it}\\
\texttt{S: ok let me look into some options for you}\\
\texttt{S: api\_call british london eight expensive}\\
U: instead could it be in madrid\\
\texttt{S: sure is there anything else to update}\\
U: no\\
\texttt{S: ok let me look into some options for you}\\
\texttt{S: api\_call british madrid eight expensive}\\
U: thank you\\
\texttt{S: you're welcome}\\
}
\end{spacing}

\noindent\textbf{dialog bAbI task 3:}
\begin{spacing}{0.9}
{\footnotesize
\noindent
\\
{
\renewcommand{\baselinestretch}{1.5}
\noindent DB Information given :\\ \\
RES\_A R\_phone RES\_A\_phone \\
RES\_A R\_cuisine french \\
RES\_A R\_address RES\_A\_address \\
RES\_A R\_location london \\
RES\_A R\_number two \\
RES\_A R\_price cheap \\
RES\_A R\_rating 2 \\
\\
RES\_B R\_phone RES\_B\_phone \\
RES\_B R\_cuisine french \\
RES\_B R\_address RES\_B\_address \\
RES\_B R\_location london \\
RES\_B R\_number two \\
RES\_B R\_price cheap \\
RES\_B R\_rating 4 \\
\\
RES\_C R\_phone RES\_C\_phone \\
RES\_C R\_cuisine french \\
RES\_C R\_address RES\_C\_address \\
RES\_C R\_location london \\
RES\_C R\_number two \\
RES\_C R\_price cheap \\
RES\_C R\_rating 7 \\
}
\\
\noindent U: hello\\
\texttt{S: hello what can i help you with today}\\
U: can you make a restaurant reservation with french food for two in london in a cheap price range\\
\texttt{S: i'm on it}\\
\texttt{S: ok let me look into some options for you}\\
\texttt{S: what do you think of this option: RES\_C}\\
U: no this does not work for me\\
\texttt{S: sure let me find an other option for you}\\
\texttt{S: what do you think of this option: RES\_B}\\
U: do you have something else\\
\texttt{S: sure let me find an other option for you}\\
\texttt{S: what do you think of this option: RES\_A}\\
U: i love that\\
\texttt{S: great let me do the reservation}\\
}
\end{spacing}

\noindent\textbf{dialog bAbI task 4:}
\begin{spacing}{0.9}
{\footnotesize
\noindent
\\
{
\renewcommand{\baselinestretch}{1.5}
\noindent DB Information given :\\
RES\_A R\_phone RES\_A\_phone \\
RES\_A R\_cuisine spanish \\
RES\_A R\_address RES\_A\_address \\
RES\_A R\_location paris \\
RES\_A R\_number two \\
RES\_A R\_price cheap \\
RES\_A R\_rating 7 \\
}
\\
\noindent U: hello\\
\texttt{S: hello what can i help you with today}\\
U: may i have a table at RES\_A\\
\texttt{S: great let me do the reservation}\\
U: do you have its address\\
\texttt{S: here it is RES\_A\_address }\\
}
\end{spacing}

\noindent\textbf{dialog bAbI task 5:}
\begin{spacing}{0.9}
{\footnotesize
\noindent
\\
{
\renewcommand{\baselinestretch}{1.5}
\noindent U: hello\\
\texttt{S: hello what can i help you with today}\\
U: can you book a table in a cheap price range\\
\texttt{S: i'm on it}\\
\texttt{S: any preference on a type of cuisine}\\
U: with british cuisine\\
\texttt{S: where should it be}\\
U: paris please\\
\texttt{S: how many people would be in your party}\\
U: for four people please\\
\texttt{S: ok let me look into some options for you}\\
\texttt{S: api\_call british paris four cheap}\\
U: actually i would prefer in a moderate price range\\
\texttt{S: sure is there anything else to update}\\
U: no\\
\texttt{S: ok let me look into some options for you}\\
\texttt{S: api\_call british paris four moderate}\\
}
\\
\noindent DB Information given :\\ \\
RES\_A R\_phone RES\_A\_phone \\
RES\_A R\_cuisine british \\
RES\_A R\_address RES\_A\_address \\
RES\_A R\_location paris \\
RES\_A R\_number four \\
RES\_A R\_price moderate \\
RES\_A R\_rating 2 \\
\\
RES\_B R\_phone RES\_B\_phone \\
RES\_B R\_cuisine british \\
RES\_B R\_address RES\_B\_address \\
RES\_B R\_location paris \\
RES\_B R\_number four \\
RES\_B R\_price moderate \\
RES\_B R\_rating 3 \\
\\
RES\_C R\_phone RES\_C\_phone \\
RES\_C R\_cuisine british \\
RES\_C R\_address RES\_C\_address \\
RES\_C R\_location paris \\
RES\_C R\_number four \\
RES\_C R\_price moderate \\
RES\_C R\_rating 1 \\
\\
\texttt{S: what do you think of this option: RES\_B}\\
U: no i don't like that\\
\texttt{S: sure let me find an other option for you}\\
\texttt{S: what do you think of this option: RES\_A}\\
U: no this does not work for me\\
\texttt{S: sure let me find an other option for you}\\
\texttt{S: what do you think of this option: RES\_C}\\
U: i love that\\
\texttt{S: great let me do the reservation}\\
U: may i have the phone number of the restaurant\\
\texttt{S: here it is RES\_C\_phone}\\
U: you rock\\
\texttt{S: is there anything i can help you with}\\
U: no thank you\\
\texttt{S: you're welcome}
}
\end{spacing}

\end{document}